# About Norms and Causes

**Daniel Kayser, Farid Nouioua**

LIPN UMR 7030 du C.N.R.S.
Institut Galilée – Univ. Paris-Nord
93430 Villetaneuse – FRANCE
dk, nouiouaf@lipn.univ-paris13.fr

**Abstract**

Knowing the norms of a domain is crucial, but there exist no repository of norms. We propose a method to extract them from texts: texts generally do not describe a norm, but rather how a state-of-affairs differs from it. Answers about the cause of the state-of-affairs described often reveal the implicit norm. We apply this idea to the domain of driving, and validate it by designing algorithms that identify, in a text, the "basic" norms to which it refers implicitly.

## 1. Motivation

**Norms are important**

The word 'norm' refers to at least two slightly different ideas: one is a kind of idealization; in this sense, formal logic can be said to be the norm of human reasoning; the other is more related to the notion of common practice. It is this latter sense that we want to explore in this paper. As a matter of fact, knowing what will normally happen next is so important, that representing norms and reasoning on them have been identified as central issues rather early in the history of Artificial Intelligence (AI).

Normal courses of events have been described in the 1970's by frames (Minsky 1974) and scripts (Schank & Abelson 1977). Capturing the ability to derive conclusions that normally follow from premises has been the major incentive for developing, in the 1980's, non-monotonic logics e.g. (AI 1980). Norms are a topic often discussed in the framework of AI and Law, and in Multi-Agent interactions (Boman 1999, Dignum et al. 2002). Reasoning on norms is often performed in this context by means of some kind of deontic logic (McNamara, Prakken 1999). Norms play also an important role in many other domains (normal evolution of social and biological systems, normal social behavior…). Less emphasized, but just as important is the role of norms in Natural Language (NL) understanding.

**Truth-based vs. Norm-based inferences in Natural Language**

Traditional NL semantics focuses on truth conditions. A text is said to entail the propositions that hold true whenever the conditions making the text true are fulfilled. This is a very weak notion of entailment. Consider for example:

(S)  *I was driving on Main Street when the truck before me braked suddenly.*

The <u>truth-based</u> entailments of (S) contain propositions like: "*there exists a time t such that I was driving at t, and there was a truck T before me at t, and T braked at t.*"

The <u>norm-based</u> entailments contain e.g.: "*Both T and me were moving in the same direction with no other vehicle in between. The distance between us was less than 100 meters.*"

Of course, the propositions of the latter list might be false while those of the former list are necessarily true if (S) is correct. Nonetheless:
• every reader of (S) will take them at least provisionally as legitimate inferences,
• the author of (S) knows that explicit indications should be provided somewhere in the text, in order to block these inferences, if they are incorrect.

Norm-based inferences are thus as rightful and much richer than truth-based ones. According to the usual tripartition between syntax, semantics and pragmatics, they might be classified under pragmatics rather than under semantics, but no matter the label, they are extremely important. The problem is how to find the norms enabling to derive them.

**How to find the norms?**

Though from a theoretical standpoint, the tools mentioned in the introduction are still perfectible, they are developed well enough to start representing our knowledge about norms. But where is this knowledge? A similar problem arises in other areas of Knowledge Representation: there exist good frameworks (Semantic Networks, Description Logics, and so on), but filling a framework with actual data is a difficult task (e.g. the CYC project, Lenat & Guha 1990).

The data from which one can start in both cases is the huge amount of textual data now available under electronic form. But the problem is easier when the goal is to elicit so-called ontologies from such texts. Kind-of and part-of hierarchies, which are the critical ingredients of an ontology, are more or less explicitly present in dictionaries, thesauri, glossaries, etc. Now, there exists no such repository in the case of norms.

**Plan of the paper**

This paper attempts to develop a methodology to extract norms from texts concerning a specific domain. We cannot expect to do so from an automatic examination of the texts: as every reader knows them, the norms are never spelled out. However, it has often been noticed that many texts, mostly narratives, describe the discrepancies between what actually happened, and the corresponding normal sequence of events. This is of course not enough to infer the norm from the text. The idea consists in using causation as a leverage, in a way explained in section 2.

Section 3 describes the method. Section 4 presents a reified first-order logic to cope with the problem of modalities. Section 5 is devoted to inference, and section 6 illustrates the whole process by a small example.

## 2. Norms and Causes

**"Why did this happen?"**

Even if it does not seem so at first sight, the notions of cause and norm are tightly related, and we will take advantage of this relationship.

Let us first notice that, strangely enough, there is very little consensus about the nature of cause. Some philosophers consider cause to be a "real" feature of the world (e.g. Kistler 1999); others consider it as a notion invented by humans to interpret phenomena, which, intrinsically, have nothing causal (see discussion e.g. in Pearl 1996). In spite of this controversy, *causal reasoning* is uncontroversially a very important issue for AI: systems for diagnosing or predicting are among the most useful applications of our discipline. In this respect, what is worth being considered as a cause must be something we can act upon. Gravitation may well be a cause of a glass being broken; having struck it is a more useful factor.

Asking for the cause of an event potentially yields an endless list; now, in a given context, only few causes come to the mind. Mackie's well-known example (1974) of an explosion happening after someone lit a cigarette is revealing: if the event takes place at home, a sensible answer can be: "*because the gas was leaking*"; if it takes place in an oil refinery, the answer might be: "*smoking in such places causes explosions*".

This example shows that "*why did this happen*"-questions concerning abnormal events, yield answers that point to a violated norm. Hence the idea that, to elicit the norms of a given domain, we should analyze the answers to questions concerning the cause of abnormal events.

(Garcia 1998) has written programs that extract causal relations from texts. Her goal is however different than ours: she attempts to collect all causal relationship present in large amounts of text, while we want to reach the norms governing a restricted domain by a close examination of some causal links.

The domain we selected to check this hypothesis concerns car accidents, because:
• we have already studied, for different reasons, a corpus of car crash reports written by drivers after an accident (tal 1994); the reports are short texts, syntactically simple, and (unfortunately) we can get as many of them as we wish;
• they describe abnormal events occurring in a domain that is non trivial, but limited well enough to make plausible an enumeration of all of its norms;
• they are good representatives of texts requiring the reader to perform norm-based inferences in order to understand what happened.

These texts have however a disadvantage: they are often biased. The author tends to describe the accident from the point of view that minimizes his/her responsibility. But this is not a real drawback: the norms that are important to collect in order to draw inferences, are those which the drivers have in mind, not the set of norms that govern the real events. Anyway no corpus can pretend to contain unbiased descriptions of events.

**Basic and Derived Anomalies**

The violation of the well-known norm:
(N1) "*under all circumstances, one must have control over one's vehicle.*"
can explain almost every accident. However, in a text like:
*In order to avoid a child suddenly rushing on the street, I swerved and bumped into a parked vehicle.*
identifying the cause of the accident as a "*loss of control*" is misleading. As a matter of fact, the violation of (N1) is a consequence of a more imperative norm:
(N2) "*harming a person must be avoided.*"

Accordingly, "*because the driver swerved to avoid the child*" is a better explanation of the accident. As accidents are anomalies, we want to trace back their causes to an anomaly; obeying (N2) being a normal behavior, we further identify the norm:
(N3) "*persons must not rush on causeways.*"
and prefer to ascribe the cause of the accident to the violation of (N3).

Similarly, we sorted out, for each car-crash report, what seems to be the "basic" cause of the accident. This work is delicate. The aim is to extract what a normal reader understands from the circumstances related, but, as we said, the description is often biased, and a normal reader knows that it can be so; therefore, understanding the report does not entail to take every statement of it for the pure truth.

Sometimes, the writer's argumentation is so obviously a purely rhetorical game, that the reader is implicitly driven to understand it as an admission of responsibility. A difficult part of the work is thus to determine up to which point the meaning intended by the writer deviates from the literal meaning of his/her text.

Anyhow, the set of causes considered basic after all the texts have been manually processed, reveals a rather comprehensive subset of the norms of the domain.

### 3. Method

**General idea**

In order to test within our limited domain of study, car-crash reports, the ideas developed in section 2, we gathered a sample of these reports and went back, in each case as said above, to what we consider as the "basic" norm violation explaining the accident. The problem then consists in designing algorithms that copy this behavior. If the algorithm correctly identifies the cause of the accident in new reports, this will mean that the set of norms collected is reasonably complete, and it will validate our approach.

This is a hard problem, involving linguistic and knowledge representation issues, as well as reasoning mechanisms.

Linguistic issues are so difficult that, by themselves, they would justify a full project. Some of the difficulties have already been identified in a previous work on the same corpus (tal 1994). Fortunately, we can skip all contextual elements (e.g. *I was driving home*), which are easily seen to be irrelevant for finding the cause of the accident.

Nonetheless, the risk is high of being stuck in nearly insoluble linguistic problems, without ever knowing whether their resolution matters for our purpose. Therefore, instead of going "forward" from the text to the norms, we started "backwards": we assume the earlier stages to be solved, and concentrate on the last ones. This method secures that we focus only on issues that are necessary.

**Layers**

Our problem is to handle the transition from hundreds of relevant propositions found in the texts, towards an expected small number of norms. It would be foolish to attempt to solve it in one single step. We therefore split it in layers: the most external one is the mere result of a parser, filtered from the irrelevant elements of the reports. The most internal one, layer 1 also called the "kernel", consists in a very small number of predicates listed in Appendix 1.

*Hypothesis H1*: All the "basic" anomalies can be explained by means of the predicates of the kernel.

Furthermore, we assume that the representation of the text at layer $n$ can be obtained from its representation at layers $n'>n$ by means of a limited number of inference rules, each one factoring out the common features that govern a specific concern. The whole process is thus a stepwise convergence from a vast variety of situations towards a smaller number of cases.

As a rule, layer $n$ is conceived in such a way that its data comes either directly from the text, or indirectly from layers $n' \bullet n$. This constraint is not sufficient to specify totally the layers; nor does it yield a total order on them. But determining exactly the boundary between layers is not critical: what is important is that each layer handles a limited number of rules, for the reasoning to remain tractable, and to avoid cycles between layers.

Currently, we specified three layers: layer 1 (the kernel) contains predicates which do not need to be further analyzed; layer 2 uses data concerning priorities, visibility, lanes, obstacles, and miscellaneous causes of loss of control; layer 3 (under construction) deals, among other concerns, with reasoning about positions of vehicles.

### 4. Time and modalities

An important issue, which remains to be dealt with, is the connection between the grammatical tenses, found in the text, and the phenomenal time (De Glas & Desclés 1996). But in order to find a satisfactory solution, we must know precisely what time needs be represented.

Several times play a role in our problem:
(i) the (linear) time of the reader: propositions in the text are ordered, but this order does not necessarily reflect the sequence of the events accounted for;
(ii) the (linear) time of the events;
(iii) the (branching) time considered by the agents: each agent indeed knows that only one future will come true, but his/her actions are explainable only by considering the possibility of several of them, among which (s)he tries to eliminate the undesirable ones.

*Hypothesis H2*: Our goal requires only the explicit representation of (ii).

*H2* is a rather strong hypothesis, since our texts abound with lexical items like "*avoid*", "*prepare*", "*expect*" (and their negation) that evoke unrealized futures. However, the author generally makes use of them for argumentation purposes, which provide no significant help in finding the causes of the accident. Notice that the hypothesis does not consider the unrealized futures as of no import, but only that these futures do not need explicit representation.

**About anomalies**

*Hypothesis H3*: *Basic* anomalies can be represented under two formats; either an agent had to do some action *a*, had the ability to do *a*, and did not *a*; or an external factor that could not reasonably be foreseen explains the accident. The *derived* anomalies are those where an agent should have done an action *a*, but, due to a basic anomaly, was not in position to do it.

According to *H3*, we need to reason on propositions of the form: MUST-DO *a* and ABLE-TO-DO *a*. They look like modals, and indeed they are; the former is clearly a kind of necessity, and the latter, a kind of possibility, but they do not obey the usual duality relationship. For one thing, an agent must not do ¬*a* every time (s)he is not able to do *a*.

## States and accessibility between states

A modal account of the duties and abilities of agents can be given by means of a possible world semantics. The accessibility between states has clearly a temporal flavor. Yet, representing every time point of a sequence of events is unnecessary; in fact, the reports look like a sequence of pictures, rather than like a film, but this metaphor is not fully adequate, since the "pictures" may use predicates that are dynamic in nature. So, a state is not characterized by the propositions which are true at a given time point, but rather by those remaining true during a given interval.

The problem is to determine the intervals. Our policy is to merge into a single state the action with its resulting state, whenever no change in modality occurs, i.e. what the agent MUST-DO and is ABLE-TO-DO remains the same once the action has been performed. On the contrary, the decision of an agent to do (or not to do) an action takes place in a state that strictly precedes the state where the action is performed.

It ensues that anomalies are found in transitions, not in states. As the usual syntax of modal logics is not well adapted for this situation, we find more convenient to represent modalities as first-order predicates in a reified logic. The two formats of *H3* are thus expressed by the formulas (*p* is the name of a predicate, Ag, the name of an agent):

(F) MUST-DO(*p*,Ag,t) ∧ ABLE-TO-DO(*p*,Ag,t) ∧ ¬HOLDS(*p*,Ag,t+1) → B-An (Basic-Anomaly)

(F') ABNORMAL-PERTURBATION(*p*,Ag,t) → B-An

The price to pay is that predicates *p* are reified; as a consequence, for representing the negation of *p*, we have to introduce a constant *not-p* and to explicit obvious facts: (∀*p*,Ag,t) HOLDS(*p*,Ag,t) ↔ ¬HOLDS(*not-p*,Ag,t) which are given for free in usual logic. Practically, this constraint is not very cumbersome. It would be more tedious if we had to reason on conjunctions or disjunctions, e.g. HOLDS(*p-and-q*,Ag,t); we never met such needs.

In order to represent scriptal unfoldings of events, we introduce a third pseudo-modal predicate: NORMALLY(*p*, Ag, t).

Pseudo-modal predicates refer implicitly to a set of accessibility relations between states, with respect to which they are actually equivalent to kripkean necessities and possibilities. For instance, NORMALLY refers to "normal" transitions. We do not develop further this aspect here.

Whereas several accessible (future) states are meaningful in most cases, *H2* says that only members of a totally ordered sequence need actually be present to reveal the basic anomalies. States are thus represented as integers, yielding a simplified version of the notion of chronicle (McDermott 1982).

## 5. Inference rules

Introducing "normal" transitions naturally leads to using non-monotonic inference rules. We use a fragment of Reiter's default rules (1980). This choice is motivated by reasons of clarity, but as defaults easily translate into auto-epistemic logic (Konolige 1988, Denecker et al. 2003), we can take advantage of several existing deductive systems. We write A : B and A : B[C] as shorthands for respectively the normal default A:B/B and the semi-normal default A:(B∧C)/B.

The basic default rule is:
NORMALLY(*p*,Ag,t) : HOLDS(*p*,Ag,t+1),
i.e. the transitions (t,t+1) in the actual unfolding of states are normal ones. Of course, since our texts report on accidents, there must be at least one exception, i.e. one actual transition that is abnormal.

In order to avoid an uncontrolled proliferation in the number of extensions, we appeal to defaults only when we have strong reasons to believe that a report could imply an exception to the rule we are expressing. In all other cases, even when exceptions are conceivable, we use material implications. Appendix 2 displays a sample of predicates and rules belonging to layer 2.

Each layer contains a small number of facts: as we said, the most external level is the output of a parser, filtered out by all elements which obviously do not resort to causal reasoning; the inner level is progressively built by the rules; as their right side is about the same length as their left side, no explosion in the number of facts is to fear. Each layer *n* is first saturated by means of rules internal to *n*; then it starts the production of facts belonging to layers • *n-1*. The inference engine stops as soon as rules (F) or (F') produces the atom B-An (basic-anomaly).

## 6. Example

Space limitations prevent us from showing but a very simple example. A significant minority of reports have this level of simplicity, and require only to reason at layers 1 and 2. But a majority of them are far more complex.

Our example is text B21 of our corpus, which reads:

> Nous nous sommes arrêtés pour laisser passer un véhicule de pompiers, la voiture qui nous suivait nous a alors heurtés.
> *We stopped to let a vehicle of firemen through; the car following us then bumped on us.*

Assuming as said above (§3), that the linguistic issues are correctly handled, we start with three states 0, 1 and 2; the initial state 0 is the same for all texts; HOLDS(Stops, A, 1) characterizes state 1; the reason of this event, the firemen, is purely contextual: we thus omit it. HOLDS(Crash, A, B, 2) and HOLDS(Is_follower, B, A, 2) characterize state 2.

The reader probably notices that HOLDS gets a varying arity (3 or 4). This is clearly forbidden in first-order logic. We present it that way for clarity [actually HOLDS is ternary and, when needed, a binary function combines together the extra-arguments. The actual expression is thus: HOLDS(COMBINE (Is_follower, A), B, t)]. The same trick is used for the pseudo-modal predicates.

Some predicates are declared static, and are endowed with forward default persistence, i.e.
STATIC(p) $\wedge$ HOLDS(p, Ag, t) : HOLDS(p, Ag, t+1)

This assumption is usual (McDermott 1982). Here, and in several other texts, we need also a kind of abductive reasoning entailing backward persistence. Being static is not enough for being backward persistent, so we declare which predicates have this feature on a case-by-case basis. Here, we do have:

HOLDS(Is_follower, Ag, Ag', t) : HOLDS(Is_follower, Ag, Ag', t-1)

This default yields: HOLDS(Is_follower, B, A, 1). Another rule tells:

($\forall$ Ag,Ag',t) HOLDS(Crash, Ag, Ag', t) $\rightarrow$ $\neg$ HOLDS(Stops, Ag', t)

i.e. that whenever Ag' bumps into Ag at time t, Ag' did not stop at time t. We thus get: $\neg$ HOLDS (Stops, B, 2).

Norm (N1) (see §2) translates into the rule:

($\forall$ Ag,t) MUST-DO (Control, Ag, t)

Agents are expected to comply with their duties, i.e.:

($\forall$ p,Ag,t) MUST-DO(p, Ag, t) $\rightarrow$ NORMALLY(p, Ag, t)

and the fact that normal events do normally happen is rendered by the normal default:

NORMALLY(p, Ag, t) : HOLDS (p, Ag, t+1)

These rules, with t=0, give: HOLDS (Control, B, 1).

We also have:

HOLDS(Is_follower, Ag', Ag, t) $\wedge$ HOLDS(Stops, Ag, t) : MUST-DO (Stops, Ag', t) [HOLDS(Control, Ag', t)]
which means that if Ag' follows Ag, and Ag stops, then Ag' must stop too, unless Ag' is not under control.

This rule provides: MUST-DO (Stops, B, 1). Finally,

($\forall$ Ag,t) HOLDS(Control, Ag, t) $\rightarrow$ ABLE-TO-DO (Stops, Ag, t)

i.e. for a vehicle, being under control implies being able to stop. This rule gives us ABLE-TO-DO (Stops, B, 1) which completes the premises of (F) (§4) with $p$ = Stops, Ag = B, t = 1, and allows to derive B-An.

This derivation stops the process. We are able to answer the question "*Why did the accident happen?*" and the answer, provided a simple NL generator is written, is: "*because vehicle B did not stop at state 2*"

## 7. Perspectives and Conclusion

### Current state and short term perspective

We have analyzed by hand a set of 73 reports. In each case, we have identified the basic anomaly and the sequence of states that is needed. We have identified the first three layers (layers 1 and 2 are illustrated in the Appendix).

We have written 74 rules and defaults to handle layers 1 and 2. Analyzing new texts will certainly show the need for new rules, or for generalizing existing ones, but we are fairly confident that the size of the whole enterprise remains manageable: at worst, a few hundreds of rules should be necessary.

The derivations are currently performed by hand. We intend to complete shortly the other layers, and to validate the approach by testing them on fresh reports.

### Longer term perspective and conclusion

There are a number of deduction engines working with various subsets of non-monotonic logics. We plan to switch as soon as possible from manual to automatic deductions. As several taggers and parsers are available for French, we will test the possibility of taking the output of one of them as the input (external layer) of our system.

But the important issue is not the success or failure of getting this work done automatically. If we have good reasons to believe that we have extracted a very small number of norms, and that a relatively small number of rules is enough to find which norms are violated, we will start to apply a similar methodology to other domains.

Asking experts what they perceive to be causes of anomalies, seems to be a good way to extract the norms in many domains. And representing the norms is of paramount importance to extend the inference capabilities beyond what is warranted by truth-conditional semantics. Moreover, being able to classify texts by the norms they are referring to, might open interesting tracks for indexing documents.

**Acknowledgment.** This work is being done in the framework of the franco-algerian project " modèles d'action " (no. 02 MDU 542) funded by the Comité Mixte d' Évaluation et de Prospective (C.M.E.P.). The authors are indebted to Françoise Gayral, François Lévy, and Adeline Nazarenko for helpful remarks on a previous version of this text.

## Appendix 1: the "kernel"

Layer 1 contains 7 reified predicates. 5 of them have a pair <vehicle, state> as arguments: Stops, Starts, Runs_slowly, Runs_backwards, Control; the last one expresses that the driver controls the vehicle during the time interval corresponding to the given state. The last two predicates are: Changes_speed (first argument is '+' or '-' depending on whether the driver speeds up or brakes, the two remaining arguments as before) and Disruptive_Factor (vehicle, name_of_factor, state).

## Appendix 2: sample of predicates and rules of layer 2

Parked, Bend, Mistaken_Command, Slippery_Road are among the binary predicates of layer 2. Crash, Visible, Obstacle, Same_File, Is_follower are ternary predicates (two vehicles and one state, e.g. Ag is visible for Ag' at state t).

A few rules connecting these predicates to one another, or inferring predicates of the kernel are given in the text (§ 6). We display here other rules, of a different flavor, to give a more comprehensive idea:

HOLDS(Bend, Ag, t) $\wedge$ ¬HOLDS(Control, Ag, t+1) : MUST-DO (Runs_slowly, Ag, t) $\wedge$ ¬HOLDS(Runs_slowly, Ag,t+1)

This default reflects an abduction: if Ag was in a bend at state t, and lost control at state t+1, it is likely that Ag had to slow down and did not do so. In order for (F) to apply, and to solve the case by deriving B-An, we must check that Ag was able to slow down, and this is the reason for the rule:

($\forall$ Ag,t) HOLDS(Control, Ag, t) $\rightarrow$ ABLE-TO-DO (Runs_slowly, Ag,t)

This next rule is hopefully self-understandable; as the previous ones, it connects layer 2 (Is_follower) with the kernel (Runs_slowly):

HOLDS(Is_follower, Ag, Ag', t) $\wedge$ HOLDS(Runs_slowly, Ag', t) $\rightarrow$ MUST-DO (Runs_slowly, Ag, t)

Finally, we show a rule internal to layer 2:

HOLDS(Same_File, Ag, Ag', t) $\wedge$ HOLDS(Crash, Ag, Ag', t) : HOLDS(Is_follower, Ag', Ag, t -1)

It captures the following abductive reasoning: if Ag' bumps into Ag and both are in the same file, it is most likely that Ag' was the follower of Ag in that file.
The whole set of predicates and rules of layers 1 and 2 can be found in (Nouioua, 2003).